\documentclass[letterpaper]{article} 
\usepackage{amsmath}
\usepackage{aaai25}  
\usepackage{times}  
\usepackage{helvet}  
\usepackage{courier}  
\usepackage[hyphens]{url}  
\usepackage{graphicx} 
\usepackage{natbib}  
\usepackage{caption} 
\usepackage{algorithm}
\usepackage{algorithmic}

\usepackage{newfloat}
\usepackage{listings}
\DeclareCaptionStyle{ruled}{labelfont=normalfont,labelsep=colon,strut=off} 
\floatstyle{ruled}
\newfloat{listing}{tb}{lst}{}
\floatname{listing}{Listing}
\title{Network Inversion of Convolutional Neural Nets}
\author{
   Pirzada Suhail,
    Amit Sethi\\
}
\affiliations{
    IIT Bombay\\

    Mumbai, IN\\
    psuhail, asethi@iitb.ac.in
}

\usepackage{bibentry}

\begin{document}

\maketitle

\begin{abstract}
Neural networks have emerged as powerful tools across various applications, yet their decision-making process often remains opaque, leading to them being perceived as "black boxes."  This opacity raises concerns about their interpretability and reliability, especially in safety-critical scenarios. Network inversion techniques offer a solution by allowing us to peek inside these black boxes, revealing the features and patterns learned by the networks behind their decision-making processes and thereby provide valuable insights into how neural networks arrive at their conclusions, making them more interpretable and trustworthy. This paper presents a simple yet effective approach to network inversion using a meticulously conditioned generator that learns the data distribution in the input space of the trained neural network, enabling the reconstruction of inputs that would most likely lead to the desired outputs. To capture the diversity in the input space for a given output, instead of simply revealing the conditioning labels to the generator, we encode the conditioning label information into vectors and intermediate matrices and further minimize the cosine similarity between features of the generated images.

\end{abstract}

%

\section{Introduction}
Neural networks have become indispensable in a wide array of applications, ranging from image recognition and natural language processing to autonomous driving and medical diagnostics. Despite their remarkable performance, the decision-making processes within these networks often remain elusive, earning them the moniker "black boxes." This opacity poses significant challenges, particularly in scenarios where interpretability and reliability are paramount, such as in safety-critical applications. Network inversion offers a solution by allowing us to inspect and reconstruct inputs that trigger specific outputs. By inverting the network, we can reconstruct inputs that are likely to produce specific outputs, thereby gaining insights into the network’s learned data distribution and feature extraction processes.

Network Inversion has been studied in different cases as in \citep{SAAD200778},\citep{784232}, \citep{Wong2017NeuralNI}, \citep{10.1145/3319535.3354261},\citep{NEURIPS2020_373e4c5d} and \citep{ansari2022autoinverseuncertaintyawareinversion}. Early research on inversion for multi-layer perceptrons in \citep{784232}, \citep{KINDERMANN1990277},\citep{SAAD200778}  derived from the backpropagation algorithm, demonstrates the utility of this method in applications like digit recognition. Subsequently multi-element evolutionary inversion procedures were introduced, that unlike gradient descent stand out for their ability to simultaneously discover multiple inversion points. Recent work by \citep{liu2022landscapelearningneuralnetwork} proposes learning a loss landscape where gradient descent becomes efficient, thus significantly improving the speed and stability of the inversion process. Later in \citep{suhail2024network} Network Inversion is performed by encoding the neural network into a Conjuctive Normal Form (CNF) Propositinal Formula and then looking for satisfying assignments to the constrained CNF formula using SAT Solvers.

In this paper, we present a simple yet effective approach to network inversion that aims to strike a balance between computational efficiency and the diversity of generated inputs by using a carefully conditioned generator trained to learn the data distribution in the input space of a classifier. To ensure the generator learns a diverse set of inputs, we alter the conditioning from simple labels to vectors and matrices that encode the label information within and discourage easy shortcut solutions. This diversity is further reinforced through the application of heavy dropout during the generation process, specifically during up-convolution, and by minimizing the cosine similarity between the features of the generated images as returned by the classifier which helps in achieving a more varied and representative set of generated images, each corresponding to different conditioning vectors. By revealing the hidden patterns and features that influence network predictions, we gain a more comprehensive understanding of neural network behavior which is crucial in improving their interpretability.

\section{Methodology}

Our approach to Network Inversion uses a conditioned generator that learns the data distributions in the input space of the trained classifier by simple modification of the training objectives as shown in Figure \ref{fig1}. Inversion is performed on a classifier which includes convolution and fully connected layers, appropriate to the classification task along with Leaky-ReLU layers\citep{xu2015empiricalevaluationrectifiedactivations} and Dropout layers \citep{JMLR:v15:srivastava14a}.

\begin{figure}[H]
\centering
\includegraphics[width=0.48\textwidth]{algo} 
\caption{Schematic Representation of the Inversion Process}
\label{fig1}
\end{figure}
The images in the input space are generated by a conditioned generator that builds up from a latent vector through up-convolution operations to the required size. While generators are typically conditioned on an embedding learned from a label for generative tasks, we observe its ineffectiveness for network inversion. Instead, we propose a more intense conditioning mechanism using vectors by encoding the label information into an n-dimensional vector for an n-class classification task that are randomly generated from a normal distribution and soft-maxed to represent the input distribution for the generated images. A linear layer maps the n-dimensional vector to the hidden dimensions of the generator before concatenation with the latent vector. The arg-max index of the soft-maxed vectors serves as the conditioning label in the cross-entropy loss function, promoting greater diversity in the generated images by varying the arg-max element intensity. The generator includes two up-sampling stages of the linear vector, the vector is upsampled into nxn dimensions followed by concatenation with an nxn hot matrix and further upsampling to the image size. The nxn hot matrix has the elements at arg-max index across both rows and columns set to 1 while others to 0.

The main objective of Network Inversion is to generate images that when passed through the classifier will elicit the same label as the generator was conditioned to. Achieving this objective through a straightforward cross-entropy loss between the conditioning label and the classifier’s output leads to mode collapse. Hence we propose Inversion loss \( \mathcal{L}_{\text{Inv}} \) as a collection of losses defined as:

\[
\mathcal{L}_{\text{Inv}} = \alpha \cdot \mathcal{L}_{\text{KL}} + \beta \cdot \mathcal{L}_{\text{CE}} + \gamma \cdot \mathcal{L}_{\text{Cosine}}
\]

Here, \( \mathcal{L}_{\text{KL}} \) is the KL Divergence loss, \( \mathcal{L}_{\text{CE}} \) is the Cross Entropy loss, and \( \mathcal{L}_{\text{Cosine}} \) is the Cosine Similarity loss. The hyperparameters \( \alpha, \beta, \gamma \) control the contribution of each individual loss term. They are defined as:
\[
\mathcal{L}_{\text{KL}} = \sum_{i} P(i) \log \frac{P(i)}{Q(i)}, \quad \mathcal{L}_{\text{CE}} = -\sum_{i} y_{i} \log(\hat{y}_{i}), \quad 
\]

\[
\mathcal{L}_{\text{Cosine}} = \frac{1}{N(N-1)} \sum_{i \neq j} \cos(\theta_{ij})
\]

where \( \mathcal{L}_{\text{KL}} \) represents the KL Divergence between the input distribution \( P \) and the output distribution \( Q \), \( y_{i} \) is the set encoded label, \( \hat{y}_{i} \) is the predicted label from the classifier, and \( \cos(\theta_{ij}) \) is the cosine similarity between the features of generated images \( i \) and \( j \) in a batch of \( N \).
\\

\section{Results}
The Network Inversion Technique was evaluated on classifiers trained on MNIST, FashionMNIST, SVHN and CIFAR10 datasets by training a generator to produce images that, when passed through a classifier, elicit the desired labels. While the generator is based on Vector-Matrix Conditioning followed by multiple layers of transposed convolutions, batch normalization \citep{pmlr-v37-ioffe15} and dropout layers \citep{JMLR:v15:srivastava14a} to encourage diversity in the generated images. 

\begin{figure}[H]
\centering
\includegraphics[width=0.46\textwidth]{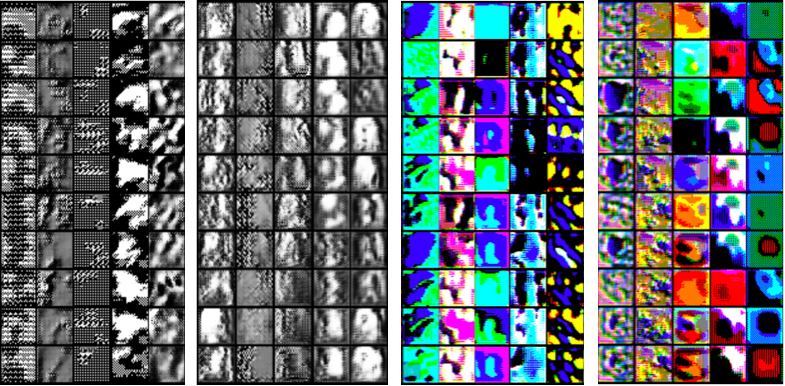}
\caption{Inverted Samples for MNIST, FashionMNIST, SVHN and CIFAR10. Each row represents a different class.}
\label{fig2}
\end{figure}

The resulting generated images are visualized to assess the quality and diversity of the generated samples in Figure \ref{fig2} for all 10 classes of MNIST and FashionMNIST, SVHN and CIFAR10 respectively. While each row corresponds to a different class each column corresponds to a different generator and as can be observed the images within each row represent the diversity of samples generated for that class. It is also observed that high weightage to cosine similarity increases both the inter-class and the intra-class diversity in the generated samples of a single generator.
Network inversion may occasionally generate samples that resemble the training data. However, to further encourage the generation of training-like data, we exploit the robustness properties of the classifiers. Typically, classifiers are not very robust to perturbations in randomly generated images but they exhibit some degree of robustness around the training examples. Consequently, by leveraging this robustness, network inversion can be guided to generate and reconstruct data that resembles the training set by encouraging the generator to produce robust inverted samples such that the perturbations of the generated images within a certain bound will still result in the same desired label.
\section{Conclusion \& Future Work}
This paper introduced a novel approach to network inversion, utilizing a single vector-matrix conditioned generator to generate a diverse set of inputs with desired output labels. 

Future work will aim to quantify the aspects of the inversion technique and explore its potential in interpretability using sparse auto-encoders on the distinct set of generated images and in training data reconstruction by encouraging the generation of robust inverted samples.
\appendix

\bibliography{aaai25}

\end{document}